ORIGINAL RESEARCH PAPER

# Using spatio-temporal deep learning for forecasting demand and supply-demand gap in ride-hailing system with anonymised spatial adjacency information

Md. Hishamur Rahman[1] | Shakil Mohammad Rifaat[2]

[1] Department of Civil Engineering, International University of Business Agriculture and Technology, Dhaka, Bangladesh

[2] Department of Civil and Environmental Engineering, Islamic University of Technology, Gazipur, Bangladesh
(Email: smrifaat@iut-dhaka.edu)

**Correspondence**
Md. Hishamur Rahman, Department of Civil Engineering, International University of Business Agriculture and Technology, Dhaka, Bangladesh.
Email: hishamur@iubat.edu

**Funding information**
Miyan Research Institute, International University of Business Agriculture and Technology.

**Abstract**
To reduce passenger waiting time and driver search friction, ride-hailing companies need to accurately forecast spatio-temporal demand and supply-demand gap. However, due to spatio-temporal dependencies pertaining to demand and supply-demand gap in a ride-hailing system, making accurate forecasts for both demand and supply-demand gap is a difficult task. Furthermore, due to confidentiality and privacy issues, ride-hailing data are sometimes released to the researchers by removing spatial adjacency information of the zones, which hinders the detection of spatio-temporal dependencies. To that end, a novel spatio-temporal deep learning architecture is proposed in this paper for forecasting demand and supply-demand gap in a ride-hailing system with anonymised spatial adjacency information, which integrates feature importance layer with a spatio-temporal deep learning architecture containing 1D convolutional neural network (CNN) and zone-distributed independently recurrent neural network (IndRNN). The developed architecture is tested with real-world datasets of Didi Chuxing, which shows that the models based on the proposed architecture can outperform conventional time-series models (e.g. ARIMA) and machine learning models (e.g. gradient boosting machine, distributed random forest, generalized linear model, artificial neural network). Additionally, the feature importance layer provides an interpretation of the model by revealing the contribution of the input features utilized in prediction.

## 1 | INTRODUCTION

Ride-hailing is recognized as a disruptive urban transportation mode where registered private car owners provide on-demand rides by driving their vehicles [1]. With the advancement of mobile technologies, ride-hailing services are replacing conventional taxi services and reshaping the mode choice behaviour of passengers. In the last few years, several ride-hailing companies such as Uber, Lyft, Didi have gained increasing popularity in many cities around the world. A recent study found that around 20% of adults in an urban area are using ride-hailing services [2]. Ride-hailing services are conducted through mobile applications, which can now adaptively coordinate the dynamic matching of passengers' requests with vacant cars [3].

From a spatial-zoning perspective, understanding and improving the spatio-temporal forecasting of short-term demand and supply-demand gap is immensely important to ride-hailing companies for determining the spatial allocation of fleet vehicles and reducing the spatial disequilibrium of supply-demand. However, due to the limited availability of ride-hailing data, less focus has been given to improve the spatio-temporal forecasting of demand and supply-demand gap in a ride-hailing system. Nevertheless, due to the similarity between taxi service and ride-hailing service, the concepts of supply-demand in the taxi system can shed important light on the ride-hailing system. Supply-demand equilibrium can be achieved theoretically if three parameters become equal: the rate of passengers demanding rides, the rate of available taxis, and the matching rate [4].

However, equilibrium cannot be maintained continuously due to several factors, such as information asymmetry, fleet size, trip fare, and short-term fluctuations [4, 5]. Therefore, disequilibrium occurs from excessive supply or excessive demand, which is unfavourable for both taxi companies and ride-hailing companies, thus necessitating them to stress on the accurate forecasting of demand and supply-demand gap. On the one hand, forecasting the demand can assist the ride-hailing companies in finding hotspot demand zones. Furthermore, along with other factors (e.g. passenger waiting time, number of trips per driver, vehicle miles travelled, and demand for parking space), demand forecasting also associates in determining appropriate fleet size [6]. While on the other hand, forecasting the supply-demand gap serves as an indicator of the supply deficit/surplus of ride-hailing cars in each zone, which is found to be informative for implementing dynamic pricing strategies and incentivizing drivers as a means to deal with the disequilibrium issues [7]. Nevertheless, there exist challenges in both the forecasting tasks due to spatio-temporal dependencies [8, 9], albeit not arising from the same factors due to varying effects of spatial and temporal dependencies in demand and supply-demand gap [10, 11].

The most popular methods to detect spatio-temporal dependencies are the convolutional neural network (CNN) [13] and the recurrent neural network (RNN) [14], which are deep learning techniques with outstanding success in tasks related to computer vision and natural language processing. Earlier studies that applied deep learning for spatio-temporal forecasting generally divide the whole study area into several zones, and the historical features corresponding to the zones are utilized as inputs in the CNN or RNN. However, these methods require further considerations for spatio-temporal forecasting with ride-hailing data containing anonymised spatial adjacency information of the zones. First, previous studies on spatio-temporal forecasting with deep learning require known spatial adjacency information for processing through two-dimensional CNN, which hinders the use of two-dimensional CNN in the case of data with anonymised spatial adjacency information. However, it is possible to process input features of such data as 1D signals through 1D CNN, based on recent findings that showed the applicability of CNN to learn from scrambled images [15]. Second, although gradient vanishing/exploding problem in the RNN for capturing long-term temporal dependencies can be controlled by utilizing the independently recurrent neural network (IndRNN) [16], applying such method directly to learn temporal dependencies from spatio-temporal features is not possible due to the internal structure of the RNN.

In this paper, we put forward a spatio-temporal deep learning architecture, the feature importance integrated 1D convolution and independently recurrent network (FOCIR-Net), to deal with the spatio-temporal dependencies for forecasting demand and supply-demand gap in a ride-hailing system with anonymised spatial adjacency information. The feature importance layer utilized in the architecture, in addition to providing an interpretation of the model, emphasizes more on strong predictors and attenuates weak predictors for each zone. To the best of the authors' knowledge, this paper, for the first time, integrates feature importance with a spatio-temporal deep learning architecture and applies 1D CNN and zone-distributed IndRNN to deal with spatio-temporal dependencies in ride-hailing data with anonymised spatial adjacency information, which can be utilized to forecast both demand and supply-demand gap in a ride-hailing system. The major contributions of this study are as follows:

(1) The FOCIR-Net detects spatial dependencies from the anonymised zones in a multivariate manner through 1D CNN and temporal dependencies in an independent manner through zone-distributed IndRNN.
(2) We integrate feature importance layer with a spatio-temporal deep learning architecture, which determines spatial weighting that adapts the model to accurately forecast both demand and supply-demand gap and also indicates the contribution of the corresponding input features for spatio-temporal forecasting.
(3) We evaluate our proposed deep learning architecture with two real-world ride-hailing datasets of Didi from Beijing and Hangzhou. The comparison of the model performance against extensively tuned statistical and machine learning models demonstrated the superiority of our model.

The remaining parts of the paper are arranged as follows. Section 2 reviews the previous studies on demand and supply-demand gap in a ride-hailing system. Section 3 defines the ride-hailing demand and supply-demand gap forecasting problem along with the associated variables. Section 4 describes the methodology and mathematical explanation of the proposed FOCIR-Net. Section 5 evaluates the proposed architecture with real-world datasets of Didi and compares the result with the benchmark models. The conclusions of the paper and future recommendations are provided in Section 6.

## 2 | RELATED WORKS

Due to recent release of some ride-hailing and taxi datasets, a large number of studies on spatio-temporal forecasting of taxi/ride-hailing demand and supply-demand gap have been conducted in the last few years. In this section, the previous related works on spatio-temporal forecasting of demand and supply-demand gap are divided into two groups based on the availability of the spatial adjacency information. First, a review of the previous works with known spatial adjacency information is provided. Second, a review of the previous works with anonymised spatial adjacency information is provided.

### 2.1 | Spatio-temporal forecasting with known spatial adjacency information

Time series forecasting techniques have been utilized for taxi demand forecasting in some of the earlier studies. Li et al. [17] designed a modified autoregressive integrated moving average (ARIMA) in an attempt to capture temporal periodicities with the short-term regular temporal patterns from the univariate time series data of historical taxi demand. The mathematical decomposition of their model further demonstrated that randomness in demand is due to contextual factors (e.g. weather condition, weekday/weekend). Moreira-Matias et al. [5] developed an ensemble of ARIMA and time-varying Poisson models to explicitly deal with short-, medium-, and long-term temporal patterns in the historical taxi demand data. However, exogenous dependencies were not considered in these studies. To that end, linear regression combined with regularization was applied to features extracted through extensive feature engineering from taxi trip records and external datasets (e.g., point of interest (POI), weather condition, events information, traffic congestion) [18, 19]. These studies showed that forecasting error reduces with the inclusion of relevant additional factors. However, all of these abovementioned studies applied the same model to all spatial units without considering the spatial dependencies, which leads to unreliable demand forecast in some spatial units due to predictability heterogeneity [20].

In recent years, machine learning has been used widely by the researchers for ride-hailing demand forecasting problems because of its ability to learning complex features by processing massive datasets. Liu et al. [21] extracted important features with the least absolute shrinkage and selection operator (LASSO) that were utilized in random forest and support vector machines (SVM) for forecasting the short-term ride-hailing demand. However, these machine learning techniques do not explicitly model the spatio-temporal dependencies. To improve this limitation, hybrid modelling approaches have become useful. Wei et al. [22] processed spatially related features through an artificial neural network (ANN) and combined them with temporal features based on demand fluctuations for forecasting the ride-hailing demand through gradient boosting decision tree (GBDT). Their study successfully incorporated the spatial correlations in demand based on the similarity of neighbouring zones, however, ignoring temporal dependencies in demand. To consider spatial and temporal dependencies simultaneously, Qian et al. [23] combined boosting methods with the Gaussian conditional random field (GCRF) model, which can be used to predict ride-hailing demand by predicting demand distributions. However, spatio-temporal dependencies were modelled only based on historical demand, which limits the detection of various patterns in spatio-temporal dependencies.

Deep learning, a branch of machine learning, offers more flexibility to design sophisticated architectures for capturing spatial and temporal dependencies in the taxi/ride-hailing demand prediction. Some of the earlier studies [24, 25] applied variants of RNN, especially long short-term memory (LSTM) network [26], for capturing the temporal dependencies in taxi/ride-hailing demand forecasting. Recent studies [27, 28] processed the historical taxi/ride-hailing demand of different zones in a city as a sequence of images and the value of historical demand for a zone in a specific time interval was considered as a pixel value. By learning spatial features and correlations among the images through CNN, the model predicts the future taxi/ride-hailing demand. Ke et al. [10] took advantage of improved partitioning by developing different types of hexagon-based convolutional neural networks (H-CNN) for different types of mapping functions (square, parity, and cube), and utilized a hexagon-based ensemble technique for forecasting supply-demand gaps. However, none of these studies was able to capture both spatial and temporal dependencies simultaneously, since CNN is unable to consider temporal patterns and RNN is unable to consider spatial patterns.

To consider both spatial and temporal dependencies in a unified framework, Shi et al. [29] developed convolutional LSTM for precipitation forecasting by combining CNN and LSTM in an architecture that modified the LSTM cell by applying convolutions in the tensor-based LSTM cell. Extending from their work, some recent studies developed ensemble of convolutional LSTMs [4] and attention-based convolutional LSTM [30] for taxi/ride-hailing demand prediction task. Ke et al. [4] also developed a spatial aggregated random forest for calculating feature importance. However, the feature importance calculated with random forest is unable to reflect the feature importance perceived by their deep learning architecture since the feature importance is not learned during the training process of the deep learning architecture. Deng et al. [31] predicted ride-hailing demand forecasting by combining ride-hailing demand similarity-based graphs with local CNN and LSTM. Geng et al. [12] encoded the region-wise correlations based on predefined factors into multiple graphs, utilized them as input patterns for ride-hailing demand forecasting by learning spatial patterns through graph convolutions and temporal patterns through context gated recurrent neural network. These studies modelled spatio-temporal dependencies in spatio-temporal forecasting with known spatial adjacency information by processing input features either as pixels or graphs. However, these techniques cannot be applied directly to data with anonymised spatial adjacency information of the zones.

## 2.2 | Spatio-temporal forecasting with anonymized spatial adjacency information

Although many studies have focused on taxi/ride-hailing demand and supply-demand gap forecasting problem with known spatial adjacency information, very few studies have explored the ride-hailing demand and supply-demand gap forecasting problem with anonymised spatial adjacency information. Li et al. [32] decomposed the ride-hailing demand time series into various frequency series through wavelet analysis, a signal processing technique, and trained SVM on the decomposed signals for forecasting the future ride-hailing demand by recomposing the predicted signals. Although their model successfully captured the non-stationarity in the time series of a zone, they disregarded the spatial dependencies among the zones.

Some of the previous studies [33, 34] utilized ensemble of various machine learning algorithms for forecasting the region-wise supply-demand gaps. Wang [34] applied a multi-layer ensemble of various machine learning algorithms (e.g. support vector machine, single XGBoost, bagging XGBoost, random forest, extra trees, and AdaBoost). The ensembles of various types of models were used to ensure diversity and stability in prediction, which helped to reduce noise and overfitting. Zhang et al. [33] applied a double ensemble technique to combine different GBDT models for forecasting supply-demand gaps in data-sparse situations. Based on a deep residual network [35], Wang et al. [9] developed a deep learning architecture for forecasting the supply-demand gap of ride-hailing services. Their framework is extendible to include new features and requires less amount of feature engineering. Besides, they used embedding technique for capturing the similarity in the supply-demand gap patterns. However, these studies were unable to explicitly model the spatio-temporal dependencies during demand and supply-demand gap forecasting with anonymised spatial adjacency information.

Our work is different from all the above mentioned studies in the following two aspects. First, previous studies designed their models to deal with spatial and temporal dependencies for forecasting taxi/ride-hailing demand and supply-demand gap in a usual scenario with known spatial adjacency information, while our proposed deep learning architecture deals with spatio-temporal dependencies in ride-hailing data with anonymised spatial adjacency information. Second, the proposed architecture in this study addresses the problem of interpretability in the spatio-temporal deep learning models for spatio-temporal forecasting by utilizing spatial weighting through the feature importance layer, which was not done previously.

## 3 | PRELIMINARIES

Both ride-hailing demand and supply-demand gap are time-series forecasting tasks, where historical values of related variables can be important indicators for predicting future values. The historical values of the ride-hailing orders, the supplied quantity of ride-hailing cars, and the number of congested roads contain useful information regarding the spatio-temporal dependencies of demand and supply-demand gap. Additionally, weather condition, POI, time of day, and day of week also affects the demand and supply-demand gap of a region.

This section presents explanations of the variables and notations used in this paper along with the forecasting problem of demand and supply-demand gap.

Definition 1 (Space-time partitioning): For variable aggregation, the study area is divided into $N$ non-overlapping uniformly sized zones $P = \{1, 2, 3, \ldots, N\}$ and total days are divided into $T$ time-slots $I = \{1, 2, 3, \ldots, T\}$ of $m$ minutes interval. The rest of the variables are defined based on the space-time partitioning.

Definition 2 (Spatio-temporal variables): The features that simultaneously vary in the spatial and temporal dimension are regarded as spatio-temporal variables. The types of the spatio-temporal variables utilized in this paper are as follows:

(1) Quantity supplied: The quantity of ride-hailing cars supplied at all zones during the time-slot $t \in I$ is represented by the vector $\mathbf{S}_t \in R^N$, where $S_{t,p}$ is the total number of ride-hailing requests at zone $p \in P$ successfully matched with drivers by the ride-hailing platform, and $S_{t,p} \in [0, +\infty)$.

(2) Demand: The total number of ride-hailing requests from a zone in a time interval is referred to as the demand, which includes both successfully matched and unanswered ride-hailing requests. The demand of all zones during the time-slot $t$ is placed in the vector $\mathbf{D}_t \in R^N$. The demand at zone $p$ during the time-slot $t$ is denoted as $D_{t,p}$, where $D_{t,p} \geq S_{t,p}$.

(3) Supply-demand gap: Supply-demand gap can be both surplus or deficit of ride-hailing cars. However, the total number of unmatched ride-hailing requests from a spatial zone, i.e., deficit in a time-slot is termed as the supply-demand gap for the rest of the paper. The supply-demand gap of all zones during the time-slot $t$ is expressed as the vector $\mathbf{G}_t \in R^N$. Therefore, the supply-demand gap of a zone $p$ during the time-slot $t$ is denoted by $G_{t,p}$, where $0 \leq G_{t,p} \leq D_{t,p}$.

(4) Traffic congestion: The traffic congestion of all zones is denoted by the vector $\mathbf{TC}_t \in R^N$, where $TC_{t,p}$ represents the total number of congested roads belonging to a zone $p$ during the time-slot $t$.

Definition 3 (Temporal variables): The temporal variables include the features that vary randomly across time, but not space. For maintaining consistency of input dimensions in our proposed architecture, these variables need to be repeated across the zones by utilizing the repeating function $f_{RZ}(\cdot; N) : R^{1 \times M} \to R^{N \times M}$, where $M$ represents the number of feature categories. The following categories of temporal variables are included in this paper:

(1) Weather condition: For the time-slot $t$, the weather conditions are represented by the row vector $\mathbf{wc}_t \in R^{1 \times C}$ consisting of $C$ weather categories (e.g., sunny, rainy, snowy etc.), where each category is encoded by one-hot encoding, i.e. $wc_{c,t} \in \{0, 1\}$. The weather condition vector $\mathbf{wc}_t$ is repeated across every zone to produce the weather condition matrix for the time-slot $t$, denoted by $\mathbf{WC}_t \in R^{N \times C}$.
(2) Temperature: The temperature (in °C) for the time-slot $t$ is repeated across the zones to produce the temperature vector $\mathbf{WT}_t \in R^N$.
(3) Particulate matter: For the time-slot $t$, the atmospheric particulate matter, $PM_{2.5}$ is repeated across the zones to form the particulate matter vector $\mathbf{WP}_t \in R^N$.

Definition 4 (Context variables): The context variables are either periodic or fixed across time. For spatio-temporal forecasting, these variables are either repeated across the zones by applying the repeating function $f_{RZ}$ or repeated across the time-slots by utilizing the repeating function $f_{RT}(\cdot; T) : R^N \to R^{N \times 1 \times T}$. These are as follows:

(1) Temporal context: Following Ke et al. [4], exploratory data analysis of the trends in demand and supply-demand gap revealed two types of periodic contexts: Time-of-day and day-of-peak. A time-slot $t$ of a day belongs to one of the three 8-h time-of-day intervals: sleep (first 8-h), peak (mid-8-h), and off-peak (last 8-h), denoted by the row vector $\mathbf{cd}_t \in R^{1 \times 3}$, where each interval $i$ is one-hot encoded, i.e. $cd_{i,t} \in \{0, 1\}$. Furthermore, a time-slot $t$ falls into one of the day-of-week categories: weekday or weekend, represented by the vector $\mathbf{cw}_t \in R$, where $cw_t \in \{0, 1\}$. The temporal context vectors $\mathbf{cd}_t$ and $\mathbf{cw}_t$ are repeated across the zones to form the time-of-day matrix $\mathbf{CD}_t \in R^{N \times 3}$ and the day-of-week vector $\mathbf{CD}_t \in R^N$ respectively.

(2) Spatial context: The spatial context refers to the number of POIs across the zones, denoted by the vector $\mathbf{cp}_p \in R^N$, which is fixed across time. Therefore, it is repeated across the time-slots with the repeating function $f_{RT}$ to form the spatial context vector for the time-slot $t$, represented by the vector $\mathbf{CP}_t \in R^N$.

With the abovementioned definition of the variables, our spatio-temporal forecasting problem can be formulated as follows: Problem: Given, the historical data of spatio-temporal variables and temporal variables up to $b$th previous time-slot starting from $t - 1$, and known data of context variables at the time-slot $t$, we are required to predict the ground truth vector $\mathbf{A}_t$ at the time-slot $t$ for spatio-temporal forecasting of demand and supply-demand in ride-hailing system with anonymised spatial adjacency information.

## 4 | METHODOLOGY

This section provides a brief description of the main components (i.e. feature importance layer, 1D CNN, and zone-distributed IndRNN) as well as the proposed architecture of the FOCIR-Net.

### 4.1 | Feature importance layer

Deep learning methods involve capturing complex interaction among features through multiple intermediate layers between inputs and outputs, which makes it difficult to interpret the contribution of the input features towards prediction. To address this issue, a number of methods [36–38] based on a one-to-one linear layer, usually utilized after the input layer, were implemented in deep artificial neural networks. The main idea of a one-to-one linear layer is similar to a fully connected layer except that a neuron in a one-to-one linear layer is connected with only one input rather than all inputs. This paper utilizes similar idea of feature importance layer except that the internal structure is modified for spatio-temporal forecasting, as expressed by the following function in Equation ( (1):

$$f_{\text{Feature Importance}}(\mathbf{X}_t) = \mathbf{X}_t \circ \sigma(\mathbf{W}^{(FI)}) \quad (1)$$

The operator $\circ$ in Equation (1) indicates Hadamard product, i.e. elementwise multiplication, between the input matrix $\mathbf{X}_t \in R^{N \times F}$ for the time-slot $t$, containing $N$ zones and $F$ input features, and the outputs of the activation function $\sigma$ (e.g. linear, sigmoid, rectified linear unit (ReLU), hyperbolic tangent, etc.) for the weight matrix $\mathbf{W}^{(FI)} \in R^{N \times F}$. For providing a fair advantage to all the features, i.e. all features are considered as equally important, training of the feature importance layer is initialized with uniform weights $\mathbf{W}^{(FI)} \sim U(-\gamma, +\gamma)$, where $\gamma$ will depend on the type of activation function utilized in Equation (1).

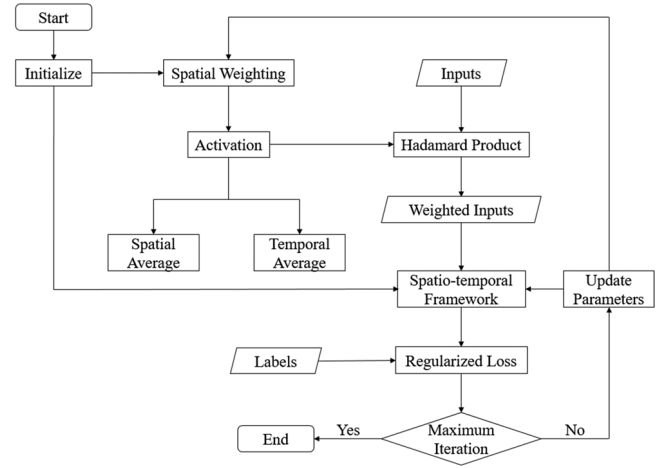

**FIGURE 1** Feature importance extraction during training

The workflow in the feature importance layer during the training process is illustrated in Figure 1. The three types of features, i.e. spatio-temporal features (demand, quantity supplied, supply-demand gap, and traffic congestion) from $t - 1$ to $t - b$, temporal features (weather, temperature, and $PM_{2.5}$) from $t - 1$ to $t - b$, and context features (time-of-day, day-of-week, POI) of the current time-slot $t$ are used as the initial inputs in the feature importance layer. These features are spatially weighted through corresponding weight matrix to produce the weighted inputs for the spatio-temporal framework consisting of 1D CNN and zone-distributed IndRNN. The feature importance outputs from the activation function are further averaged spatially and temporally for explaining feature contribution in the model from temporal and spatial perspective, respectively. It is noteworthy to mention that sigmoid is utilized as the activation function in the feature importance layer in this study. The use of sigmoid activation in the feature importance layer is decided empirically, i.e. contributed to minimum model loss in comparison to other activation functions (linear, ReLU, and tanh).

The theoretical justification of the feature importance layer utilized in this paper is similar to Borisov et al. [36]. The weight matrix is updated during the training process, assigning larger weights to more important features and smaller weights to less important features. To ensure that the feature importance layer correctly captures the contribution of the input features, the weights and biases of the subsequent layers must not become zero during the training process. This is maintained by utilizing the L2-norm of regularization for the weights and biases in the layers after the feature importance layer in the training algorithm. Furthermore, to ensure sparsity of $\mathbf{W}^{(FI)}$, the L1-norm of regularization for $\mathbf{W}^{(FI)}$ is included during the training of the architecture.

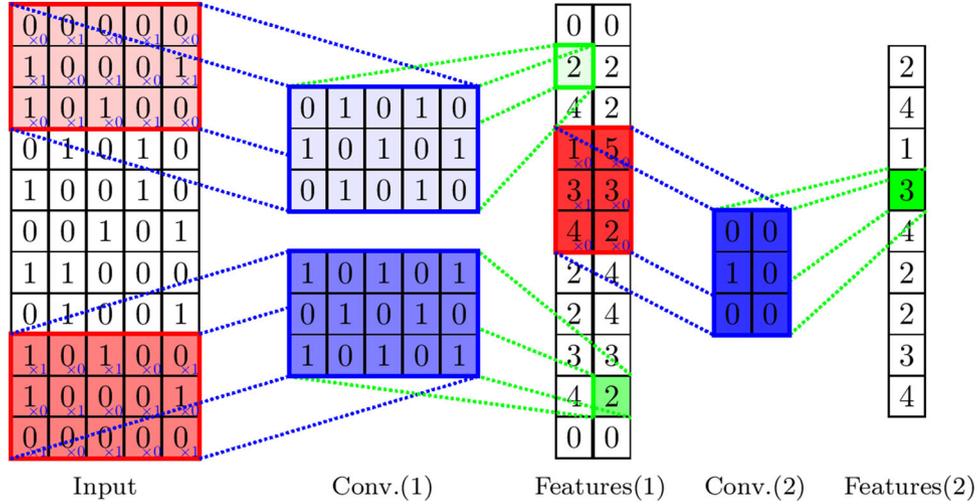

**FIGURE 2** An example 1D CNN

## 4.2 | 1D CNN

The CNN is a specialized neural network for detecting spatial dependencies by various types of filters (i.e. weights and bias) sliding over and convolving with the input. To address the spatial dependencies from anonymised zones in spatiotemporal forecasting, a 1D CNN is utilized in this paper, where the filters are slid over only across the spatial dimension (i.e. anonymised zones) of the input with spatio-temporal feature columns. To detect same kind of spatial features with a filter, parameters of a filter are shared across the zones. The output feature vector of a filter $k$ in a 1D CNN layer can be expressed as follows:

$$\mathbf{Z}^{(k)} = \sigma \left( \mathbf{X}_t * \mathbf{W}^{(k)} + \mathbf{b}^{(k)} \right) \quad (2)$$

where $*$ is the convolution operator, $\mathbf{X}_t$ is the input matrix convolved with the filter $k$, $\mathbf{Z}^{(k)} \in R^N$ refers to the output feature vector for the filter, $\mathbf{W}^{(k)} \in R^{E \times F}$ serves as the shared weight matrix of the filter with length $E$, and $\mathbf{b}^{(k)} \in R^N$ is the shared bias vector of the filter. Therefore, the 1D CNN layer with $K$ filters applied to the input matrix $\mathbf{X}_t$ can be represented by the function $f_{Conv1D} : R^{N \times F} \to R^{N \times K}$.

In this study, a 1D CNN with two layers is utilized. The first layer contains 200 filters and the second layer contains 400 filters. A stride of one is utilized for sliding the filters. Varying filter size ranging from 3 to 13 are tested for the filters to determine the best filter size. An example of the inputs and outputs in a 1D CNN with two convolutional layers are shown in Figure 2. The first convolutional layer is comprised of two filters and the second convolutional layer is comprised of a single filter. The outputs of the convolutional layers are the columns of features, which are spatial patterns from the different types of input features utilized during 1D convolutions with the filters. To keep the dimensions of inputs and outputs equal, zero paddings are used in the inputs of each convolutional layer. Although conventional CNN includes pooling layers, in order to prevent loss of spatial information, our 1D CNN avoided the pooling layer following [27, 39].

## 4.3 | Zone-distributed IndRNN

The RNN is an exclusive architecture for detecting temporal dependencies from time-series data where the features of one time-slot are correlated with the features of the previous time-slots. In comparison to the non-recurrent connections in conventional neural networks, the RNN has recurrent connections, i.e. the outputs of the hidden layer neurons from the previous time step of a sequence are utilized with the inputs of the current time step, which can be expressed by Equation (3):

$$\mathbf{h}_t = \sigma \left( \mathbf{U}\mathbf{x}_t + \mathbf{W}\mathbf{h}_{t-1} + \mathbf{b} \right) \quad (3)$$

where $\mathbf{x}_t \in R^F$ is the input vector at the current time step containing $F$ features, $\mathbf{h}_t \in R^H$ and $\mathbf{h}_{t-1} \in R^H$ refer to the output vectors of size $H$ representing hidden layer neurons of current and previous time step respectively, $\mathbf{U} \in R^{H \times F}$ and $\mathbf{W} \in R^{H \times H}$ serve as the weights for the input of the current step and the outputs of the previous step respectively, and $\mathbf{b} \in R^H$ is the bias vector.

However, repeated multiplication of the recurrent hidden layer weight matrix during training results in the vanishing/exploding gradient problem in RNN that limits the storing of long-term information, which was tackled through LSTM by including additive updates in the hidden layer [26]. Despite the modification, training of LSTM can also suffer from vanishing gradient problem due to the application of non-linear activation functions [16], and can also face exploding gradient problem due to self-multiplication of the weight matrix [40]. To overcome these problems, IndRNN was proposed by replacing the entangled recurrent connections in Equation (3) with an element wise multiplication, which can be utilized to learn temporal dependencies in spatio-temporal forecasting from the independent recurrent connections, as shown in Equation (4):

$$\mathbf{h}_t = \sigma \left( \mathbf{U}\mathbf{x}_t + \mathbf{W}^{(I)} \circ \mathbf{h}_{t-1} + \mathbf{b} \right) \quad (4)$$

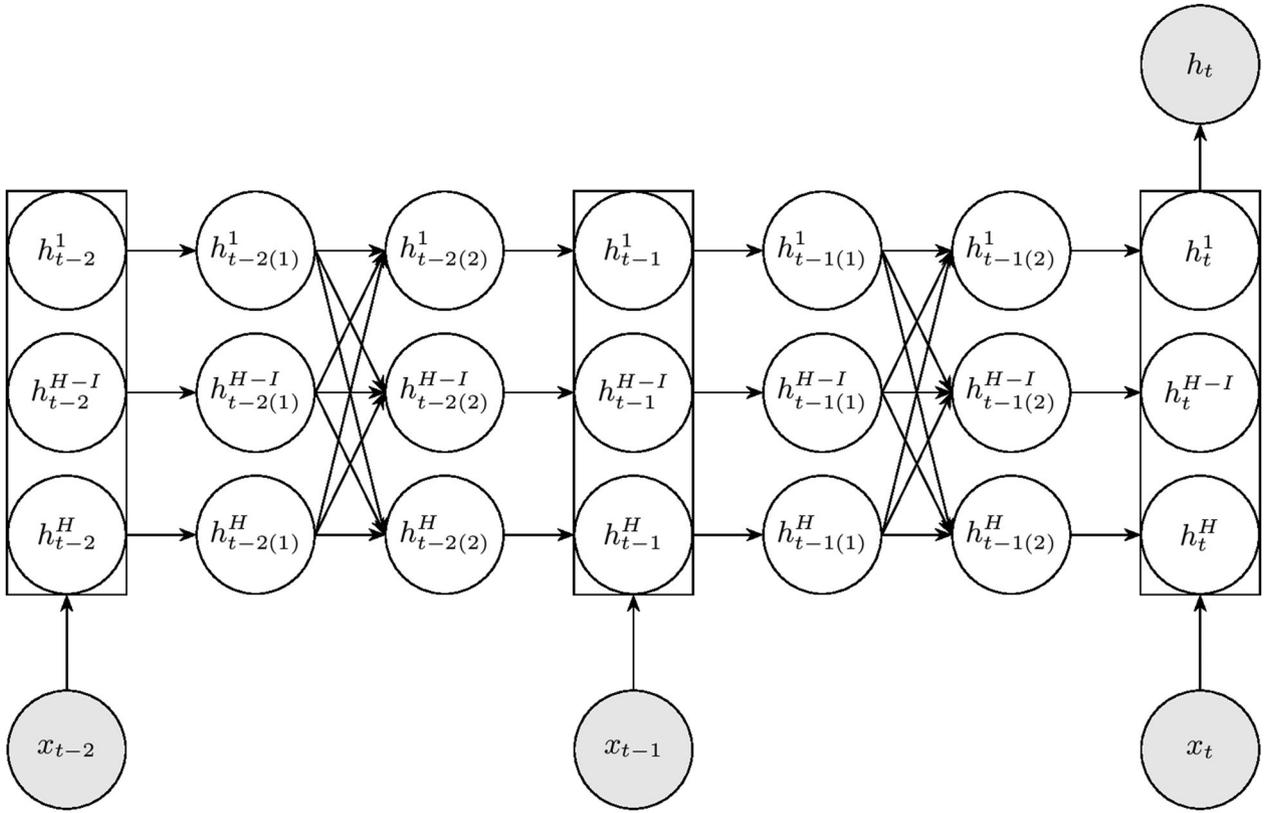

**FIGURE 3** An example IndRNN

where $\mathbf{W}^{(I)} \in R^H$ refers to the weight vector for the independently recurrent outputs of the last time step, which is regulated depending upon the activation function, i.e. ReLU/tanh following the theoretical constraints provided by [16] for preventing vanishing/exploding gradient problem during training of the IndRNN. In this study, both ReLU and tanh are tested as the activation function in the IndRNN and the one resulting in minimum loss in the FOCIR-Net is finally chosen.

An example of an unfolded IndRNN is shown in Figure 3, which has two hidden layers (each containing $H$ neurons where $h$ represents an individual neuron) processing an input sequence containing three time steps. In the first hidden layer, neurons independently receive the input and the corresponding outputs of the previous time step, which facilitates the detection of various temporal dependencies from the input. The subsequent hidden layer further explores the correlation among the independently learned temporal dependencies. The output vector of the current time step $t$ is the required prediction.

For detecting temporal dependencies in spatio-temporal forecasting, instead of a vector, the input of IndRNN at a time step is a matrix including the spatial dimension (i.e. zones), which requires the same IndRNN to be distributed across the zones independently with shared parameters. Therefore, we applied the IndRNN layer to the $N$ zones of the input matrix $\mathbf{X}_t$ by utilizing the function $f_{ZoneDistributed(IndRNN)} : f_{IndRNN}(\mathbf{x}_t) \rightarrow f_{IndRNN}(\mathbf{X}_t)$.

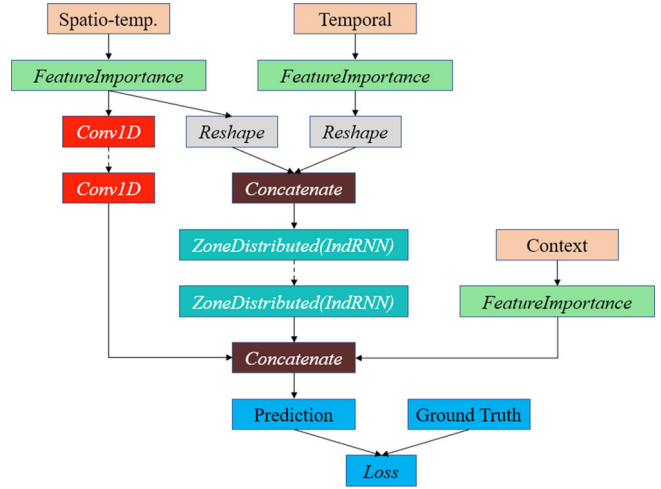

**FIGURE 4** Workflow of FOCIR-Net architecture

### 4.4 | Architecture of FOCIR-Net

The workflow of the FOCIR-Net architecture is shown in Figure 4. All the inputs are initially assigned spatial weights through the feature importance layer to learn their contribution to the prediction. The weighted spatio-temporal variables are then passed through the layers of 1D CNN to learn spatial dependencies, meanwhile, the weighted temporal variables together with the weighted spatiotemporal variables are passed through the stacked layers of zone-distributed IndRNN to learn temporal dependencies. Finally, the outputs of the 1D CNN and zone-distributed IndRNN are then combined with the weighted context features to make the final prediction $\mathbf{O}_t$. According to the problem mentioned in Section 3, the prediction target is the demand $\mathbf{D}_t$ or the supply-demand gap $\mathbf{G}_t$, which is denoted as ground truth $\mathbf{A}_t$. Techniques such as concatenating and reshaping are utilized to adapt the inputs with the dimensional requirements in the architecture. To concatenate different matrix in our architecture, the concatenation function $f_{Concatenate}$ is applied, which combines the matrices along the feature axis. For example, if $\mathbf{P} \in R^{N \times L}$ and $\mathbf{Q} \in R^{N \times M}$ are two matrix with $N$ zones containing $L$ and $M$ features, respectively, to be concatenated, then the concatenation function can be expressed as $f_{Concatenate}(\mathbf{P}; \mathbf{Q}) : (R^{N \times L}; R^{N \times M}) \rightarrow R^{N \times (L+M)}$. The reshaping function $f_{Reshape}$ is specifically utilized to adapt the weighted spatio-temporal and temporal inputs with the time-step requirement in the zone-distributed IndRNN, i.e. to convert input matrix to tensor. For example, if $\mathbf{X}_t^{(ST)} \in R^{N \times Bb}$ is the weighted spatio-temporal matrix with $N$ zones containing $B$ features (i.e. demand, quantity supplied, supply-demand gap, and traffic congestion), each taken up to $b$ previous time-steps (i.e., from $t-1$ to $t-b$), then the reshaping function can be expressed as $f_{Reshape}(\mathbf{X}_t^{(ST)}; B; b) : R^{N \times Bb} \rightarrow R^{N \times B \times b}$.

The training of the FOCIR-Net involves minimizing the mean squared error between the predicted values and the ground truth, which is achieved through the loss function $f_{Loss}$. The objective of the loss function applied in our architecture including regularization terms can be expressed as Equation (5):

$$\min_{W^{(A)}, W^{(FI)}, b} f_{Loss} = \|\mathbf{O}_t - \mathbf{A}_t\|_2^2 + \alpha \|\mathbf{W}^{(A)}\|_2^2 + \beta \|\mathbf{W}^{(FI)}\|_1 \quad (5)$$

where $\mathbf{W}^{(A)}$ refers to all parameters of the FOCIR-Net except the feature importance layer parameter $\mathbf{W}^{(FI)}$, and $\alpha, \beta$ are the regularization parameters. The L1- and L2-norm of regularization are utilized in accordance with the requirements of the feature importance layer, which also assists in avoiding overfitting issues.

## 5 | EXPERIMENTS AND DISCUSSIONS

This section presents the experiments on real-world data and the discussion of their results. Furthermore, model ablation of the proposed FOCIR-Net is also provided. Finally, the interpretations of the FOCIR-Net models are provided at the end of the section.

### 5.1 | Data description and preprocessing

Publicly released ride-hailing datasets [41] from two cities of China, Beijing and Hangzhou, are selected for the experiments in this paper. Didi divided each city into several square zones (i.e. Beijing into 66 zones and Hangzhou into 58 zones) by applying geohashing and identified each zone with a unique ID, anonymizing the adjacency information among the zones.

We utilized the Beijing dataset spanning from 1st January 2016 to 20th January 2016 and the Hangzhou dataset spanning from 23rd February 2016 to 17th March 2016. To construct time-series data for each zone, the total timespan is divided into equal interval time-slots. Considering the small length of the datasets and focusing on short-term forecasting, each day in the datasets is therefore divided into 144 time-slots of 10 min interval. Furthermore, the total time-slots in a zone are split into training, validation, and testing sets for our experiments. Around 30% of the time-slots are reserved for validation and testing, and the rest of the data are used in training the models. The Beijing dataset contains 190,080 (66 zones × 20 days × 144 time-slots) samples, of which 132,000 samples are utilized in training and 58,080 samples are used in validation and testing. The Hangzhou dataset contains 200,448 (58 zones × 24 days × 144 time-slots) samples, of which 142,448 samples are utilized in training and 58,000 samples are used in validation and testing.

The raw datasets provide four types of information: order information, traffic congestion information, weather information, and POI information. The order information and traffic congestion information are aggregated based on zones and time-slots to extract the spatio-temporal variables, i.e. demand, supply-demand gap, quantity supplied, and traffic congestion. The weather information provides temporal variables, i.e. weather, temperature, and $PM_{2.5}$. The spatial context variables are aggregated from the POI information and the temporal context variables, i.e. time-of-day and day-of-week are extracted from the order information.

Both the datasets contain information around 8.5 million ride-hailing orders. The raw order information provides anonymised information of each order including driver ID, passenger ID, and trip origin and destination geohashes in string format. Besides, date and time of the orders are also available in the datasets, which are assigned corresponding time-of-day and day-of-week values. The unfulfilled orders are marked by a 'null' driver ID, which is useful information for counting the supply-demand gaps in a zone for a time-slot. Furthermore, in order to find the demand and quantity supplied of each zone from the order information, the orders including 'null' driver IDs and orders excluding 'null' driver IDs are counted, respectively, for a timeslot. For both the datasets, around 50% of the time-slots are found to have zero supply-demand gaps, around 17% of which are due to zero demand, indicating that orders in around 33% time-slots are fully matched by the platform. The spatial distribution of the orders across zones, including the proportions of quantity supplied and supply-demand gap, is shown in Figures 5(a) and 5(b) for Beijing and Hangzhou respectively. In general, zones with higher demand tend to show a higher supply-demand gap in both datasets. Typical weekly patterns of demand and supply-demand gap from different zones of Beijing and Hangzhou are presented in Figure 6.

The zone-wise traffic congestion information for each time-slot indicates the number of roads at four different congestion

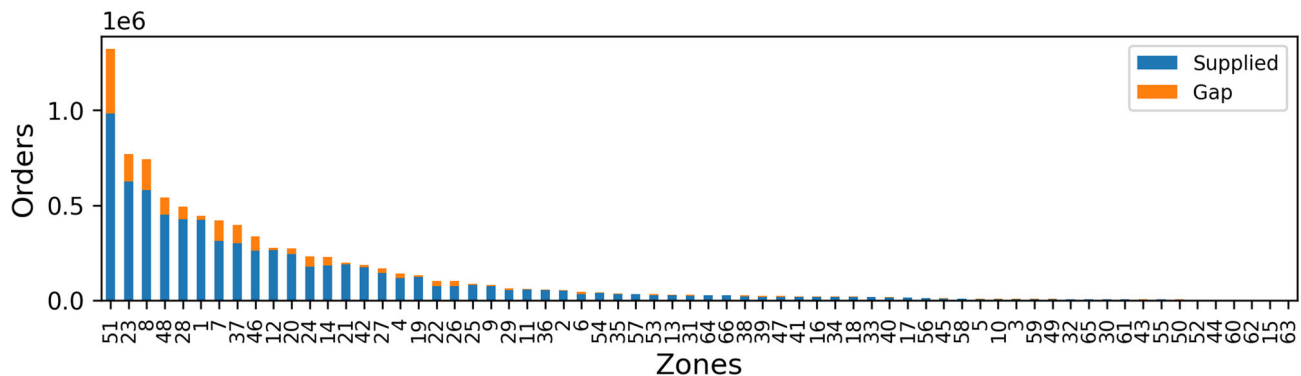

(a) Spatial distribution of orders across zones in Beijing

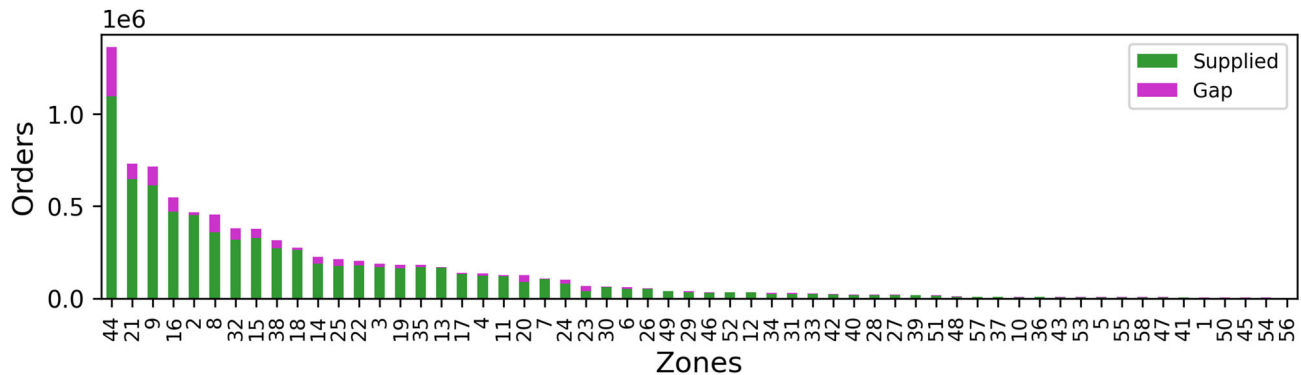

(b) Spatial distribution of orders across zones in Hangzhou

**FIGURE 5** Spatial distribution of orders across zones

levels. However, the meanings of these congestion levels are unknown. Therefore, these values are summed to get an idea of the overall traffic congestion of a zone in a time-slot.

The values of weather category, temperature, and $PM_{2.5}$ is available in the dataset at 10 min interval. These values are matched with the corresponding time-slots to get the temporal variables of a time-slot.

The zone-wise number of facilities for 24 different POI classes are available in the POI information. However, since the names of these POI classes are not available, they do not provide any meaningful information. Therefore, the number of POIs of different classes are summed to get the total number of POIs in a zone.

## 5.2 | Model evaluations

The proposed FOCIR-Net is compared against a set of benchmark algorithms and an ablation analysis of FOCIR-Net is conducted. For a fair comparison, all models utilized the same lookback window up to sixth previous time-slot. The settings of FOCIR-Net are decided by tuning of the hyperparameters. The finalized settings of the hyperparameters are presented in Table 1.

In order to assess the performance of FOCIR-Net, a number of benchmark models are considered, which are extensively tuned by utilizing automated machine learning frameworks. They are as follows:

(1) Gradient Boosting Machine (GBM): The GBM [42] is an ensemble method that is built upon several additive regression trees through the utilization of the gradient descent technique.
(2) Extreme Gradient Boosting (XGBoost): The XGBoost [43], a modified version of GBM, is a popular algorithm that has topped in many machine learning competitions. The XGBoost algorithm utilizes regularized gradient boosting to control overfitting.
(3) Random Forest (RF): The RF [44] is an ensemble method that utilizes several weak learner regression trees. The splitting of the trees is determined by finding the most appropriate discriminative threshold for a subset of features.
(4) Extremely Randomized Trees (XRT): The XRT [45] is similar to the RF except that it extracts randomly generated thresholds for each feature and the best threshold is utilized for splitting the trees.
(5) Generalized Linear Model (GLM): The GLM [46] is a generalization of the linear regression that allows any of the exponential family of distributions in the errors of the outcome variable. Since our outcome is continuous, our GLM implementation assumes Gaussian distribution.

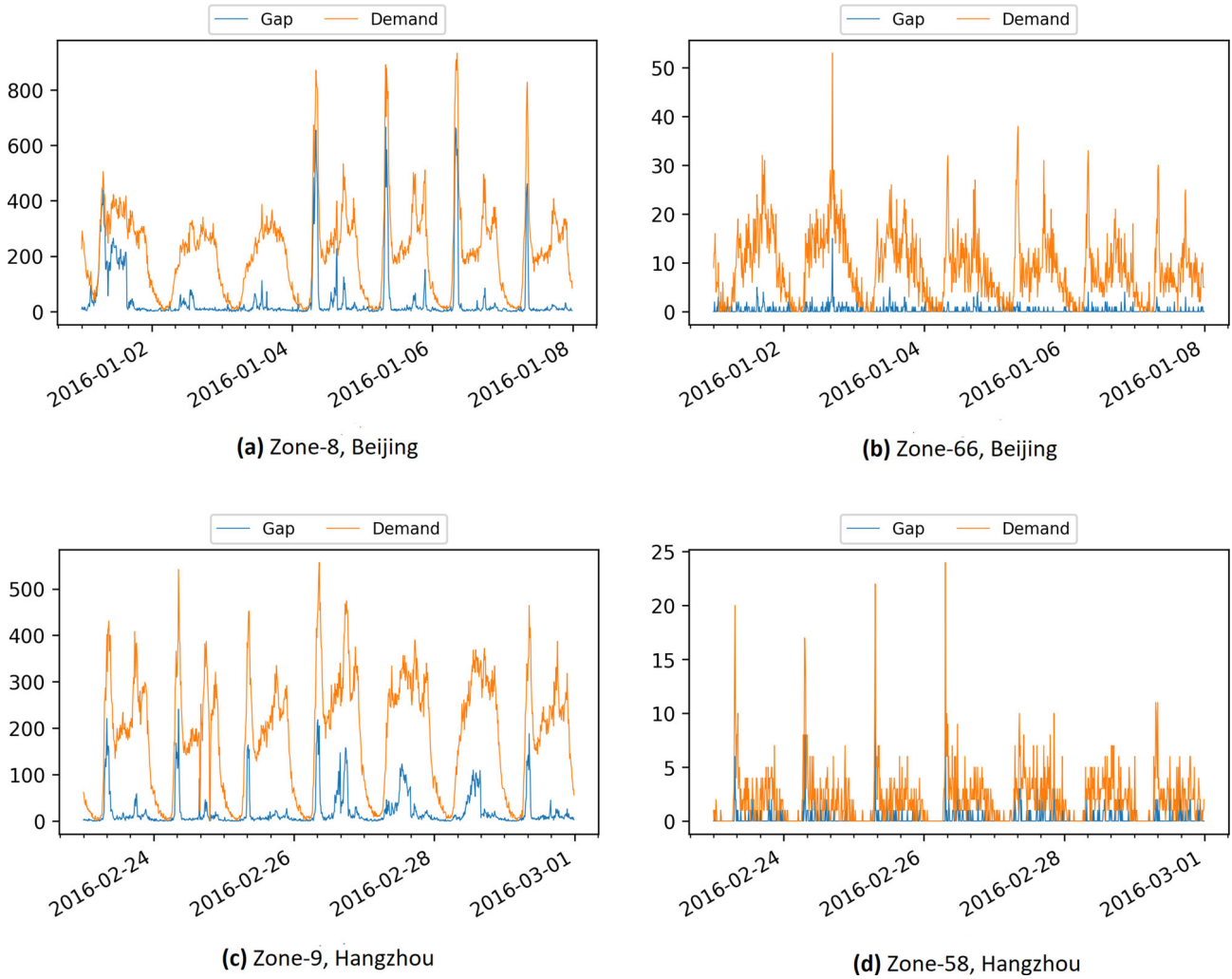

**FIGURE 6** Demand and supply-demand gap of various zones in Beijing and Hangzhou

(6) Artificial Neural Network (ANN): The ANN [48] contains a feed forward neural network with several hidden layers to learn hierarchical representation from the input features. Our ANN is trained with a mini-batch size of one.

(7) Autoregressive Integrated Moving Average (ARIMA): The ARIMA [49] is a time series forecasting method that removes trends from the data and estimates relationship and error from the past observations to forecast future values. Our implementation of ARIMA includes explanatory variables with conventional ARIMA.

It is noteworthy to mention that, all the benchmarks excluding ARIMA utilized the same input features as FOCIR-Net, whereas ARIMA utilized all input features except the categorical ones.

In addition to the abovementioned models, model and feature ablation of the FOCIR-Net is also conducted by removing different model components and feature sets. The following configurations are tested for model ablation:

(1) 1D Convolution and Independently Recurrent Network (OCIR-Net): The OCIR-Net includes all model components except the feature importance layer. The input features are directly fed into the model without any spatial weighting.

**TABLE 1** Hyperparameter settings of FOCIR-Net

| | **Hyperparameter settings** |
|---|---|
| (a) Feature Importance Layer | Weight initialization: uniform; activation: sigmoid |
| (b) 1D-CNN | Layers: 2 layers; filters: 200 in 1st layer; 400 in 2nd layer |
| | Weight initialization: uniform; activation: ReLU |
| (c) Zone-distributed IndRNN | Layers: 2 layers; hidden units: 4 units per layer |
| | Weight initialization: uniform; activation: tanh/ReLU |
| (d) Fully Connected Layer | Weight initialization: uniform; activation: ReLU |
| (e) Model Training | Optimizer: Adam [47]; learning rate: 0.001; batch size: 32 |
| | Regularization: L1 = 0.001, L2 = 0.001; early stopping: patience = 100 epochs |

**TABLE 2** Performance evaluation of FOCIR-Net and benchmark models

|  | Model | Metrics (Demand) | | | | Metrics (Supply-demand gap) | | | |
| --- | --- | --- | --- | --- | --- | --- | --- | --- | --- |
|  |  | MAE | RMSE | sMAPE | Time (s) | MAE | RMSE | sMAPE | Time (s) |
| (a) Beijing | FOCIR-Net | 6.45 | 16.77 | 0.1867 | 229 | 3.34 | 14.56 | 0.2288 | 555 |
|  | XGBoost | 6.88 | 19.07 | 0.2009 | 3.68 | 3.70 | 16.74 | 0.2952 | 11.89 |
|  | GBM | 7.00 | 18.91 | 0.2186 | 3.99 | 3.77 | 16.53 | 0.3053 | 3.43 |
|  | XRT | 7.04 | 19.43 | 0.1994 | 78.40 | 3.91 | 17.07 | 0.3045 | 31.03 |
|  | RF | 7.23 | 22.18 | 0.2006 | 79.45 | 3.90 | 16.60 | 0.3044 | 37.32 |
|  | GLM | 7.65 | 19.78 | 0.2537 | 0.44 | 4.61 | 16.80 | 0.3998 | 1.01 |
|  | ANN | 14.48 | 25.57 | 0.4250 | 33.21 | 5.34 | 19.09 | 0.4671 | 114.74 |
|  | ARIMA | 8.22 | 20.79 | 0.2152 | 262.73 | 4.57 | 17.52 | 0.3321 | 201.60 |
| (b) Hangzhou | FOCIR-Net | 6.82 | 16.34 | 0.1626 | 237 | 2.92 | 13.45 | 0.2257 | 181 |
|  | XGBoost | 7.25 | 17.35 | 0.1770 | 17.18 | 3.06 | 13.46 | 0.2873 | 14 |
|  | GBM | 7.12 | 17.09 | 0.1761 | 10.69 | 3.06 | 13.52 | 0.3018 | 4.24 |
|  | XRT | 7.19 | 17.43 | 0.1734 | 106.97 | 3.12 | 13.64 | 0.2898 | 59.59 |
|  | RF | 7.18 | 17.28 | 0.1730 | 103.09 | 3.14 | 13.80 | 0.2896 | 58.76 |
|  | GLM | 7.70 | 18.05 | 0.2070 | 0.42 | 3.44 | 14.55 | 0.3222 | 0.82 |
|  | ANN | 7.80 | 18.12 | 0.2137 | 7.67 | 3.41 | 14.07 | 0.3426 | 7.85 |
|  | ARIMA | 11.33 | 26.32 | 0.2617 | 242.73 | 7.88 | 28.60 | 0.4627 | 454.55 |

**TABLE 3** Model ablation analysis of FOCIR-Net

|  | Model | Metrics (Demand) | | | Metrics (Supply-demand gap) | | |
| --- | --- | --- | --- | --- | --- | --- | --- |
|  |  | MAE | RMSE | sMAPE | MAE | RMSE | sMAPE |
| (a) Beijing | OCIR-Net | 6.71 | 17.19 | 0.1942 | 3.38 | 14.67 | 0.2351 |
|  | FOC-Net | 6.49 | 16.96 | 0.1958 | 3.42 | 14.98 | 0.2372 |
|  | FIR-Net | 19.61 | 51.00 | 0.3116 | 7.44 | 33.13 | 0.3057 |
|  | FIN-Net | 8.57 | 23.02 | 0.2314 | 5.78 | 21.25 | 0.3379 |
|  | 1D-CNN | 6.49 | 16.86 | 0.1970 | 3.52 | 15.39 | 0.2422 |
|  | Zone-distributed IndRNN | 21.39 | 58.08 | 0.3974 | 12.33 | 41.04 | 0.3530 |
| (b) Hangzhou | OCIR-Net | 7.02 | 16.68 | 0.1718 | 2.93 | 13.59 | 0.2234 |
|  | FOC-Net | 7.04 | 16.66 | 0.1693 | 3.08 | 13.91 | 0.2298 |
|  | FIR-Net | 7.71 | 18.25 | 0.1928 | 3.61 | 15.25 | 0.2583 |
|  | FIN-Net | 7.85 | 18.34 | 0.2006 | 3.51 | 14.95 | 0.2526 |
|  | 1D-CNN | 7.23 | 17.18 | 0.1823 | 3.10 | 13.91 | 0.2319 |
|  | Zone-distributed IndRNN | 7.84 | 18.44 | 0.1987 | 3.65 | 15.20 | 0.2598 |

(2) Feature Importance Integrated 1D Convolution Network (FOC-Net): The zone-distributed IndRNN is removed in this configuration. In this model, the weighted spatio-temporal features are processed through the 1D CNN and the rest of the weighted features (i.e. temporal and context features) are concatenated with the outputs of the 1D CNN to make the final prediction.

(3) Feature Importance Integrated Independently Recurrent Network (FIR-Net): The FIR-Net excludes the 1D CNN. Only temporal dependencies are considered in this model.

(4) Feature Importance Neural Network (FIN-Net): The FIN-Net excludes both 1D CNN and zone-distributed indRNN. The spatially weighted features are passed through a fully connected neural network in this model.

(5) 1D Convolution Neural Network (1D-CNN): The feature importance layer and the zone-distributed IndRNN are removed in this configuration. In this model, the spatio-temporal features are directly processed through the 1D CNN and the rest of the features (i.e. temporal and context features) are concatenated with the outputs of the 1D CNN to make the final prediction.

**TABLE 4** Feature ablation analysis of FOCIR-Net

|   | Feature Combination | Metrics (Demand) | | | Metrics (Supply-demand gap) | | |
|---|---|---|---|---|---|---|---|
|   |   | MAE | RMSE | sMAPE | MAE | RMSE | sMAPE |
| **(a) Beijing** | Spatio-temporal + Temporal | 6.64 | 16.88 | 0.2103 | 3.32 | 14.63 | 0.2397 |
|   | Spatio-temporal + Context | 6.62 | 17.12 | 0.1970 | 3.33 | 14.64 | 0.2477 |
|   | Temporal + Context | 29.72 | 76.73 | 0.4770 | 12.69 | 45.56 | 0.4534 |
|   | Spatio-temporal | 6.61 | 17.00 | 0.2099 | 3.35 | 14.78 | 0.2655 |
|   | Temporal | 38.94 | 102.90 | 0.4324 | 16.85 | 48.69 | 0.4867 |
|   | Context | 28.95 | 77.42 | 0.5171 | 9.55 | 43.98 | 0.3939 |
| **(b) Hangzhou** | Spatio-temporal + Temporal | 7.47 | 17.75 | 0.1843 | 2.97 | 13.71 | 0.2305 |
|   | Spatio-temporal + Context | 7.05 | 16.81 | 0.1853 | 2.98 | 13.69 | 0.2503 |
|   | Temporal + Context | 20.22 | 53.58 | 0.2696 | 6.30 | 29.95 | 0.2921 |
|   | Spatio-temporal | 7.03 | 16.83 | 0.1823 | 3.02 | 13.68 | 0.2622 |
|   | Temporal | 29.25 | 66.14 | 0.3973 | 7.09 | 29.59 | 0.3494 |
|   | Context | 22.74 | 57.08 | 0.4120 | 6.46 | 26.85 | 0.3361 |

(6) Zone-distributed Independently Recurrent Network (Zone-distributed IndRNN): The Zone-distributed IndRNN excludes the feature importance layer and the 1D CNN. The reshaped spatio-temporal variables and temporal variables are combined and directly processed through the zone-distributed IndRNN and the rest of the features (i.e. temporal and context features) are concatenated with the outputs of the zone-distributed IndRNN to make the final prediction.

The following feature combinations are tested for feature ablation in the FOCIR-Net:

1. Spatio-temporal + Temporal: Context features are removed from the inputs.
2. Spatio-temporal + Context: Temporal features are removed from the inputs.
3. Temporal + Context: Spatio-temporal features are removed from the inputs.
4. Spatio-temporal: Tested with spatio-temporal features only.
5. Temporal: Tested with temporal features only.
6. Context: Tested with context features only.

The performances of the models utilized in this paper are evaluated with three metrics: mean absolute error (MAE), root mean squared error (RMSE), and symmetric mean absolute percentage error (sMAPE), which can be computed by using Equations (6)–(9):

$$MAE = \frac{1}{n} \sum_{i=1}^{n} |\mathbf{O}_i - \mathbf{A}_i| \quad (6)$$

$$RMSE = \sqrt{\frac{1}{n} \sum_{i=1}^{n} (\mathbf{O}_i - \mathbf{A}_i)^2} \quad (7)$$

$$sMAPE = \frac{1}{n} \sum_{i=1}^{n} \frac{|\mathbf{O}_i - \mathbf{A}_i|}{|\mathbf{O}_i| + |\mathbf{A}_i| + 1} \quad (8)$$

where $\mathbf{O}_i \in R^N$ and $\mathbf{A}_i \in R^N$ are the predicted vector and ground truth vector, respectively, at time-slot $i$ in the test set with $n$ time-slots and $N$ zones. Since our target value contains zero and sMAPE produces inaccurate statistics when encountered with zero, therefore, a modified sMAPE [5] is utilised.

Our proposed architecture is trained on a server with 4 Core (hyper-threaded) Xeon processor (2.30 GHz), 25 GB RAM, and a Tesla K-80 GPU. The FOCIR-Net is written in Python 3 using Keras [50] with Tensorflow [51] backend. All the benchmarks except ARIMA are implemented in H2O AutoML [52], while ARIMA is implemented in Pmdarima [53].

The performances of the FOCIR-Net and the benchmark models are reported in Table 2. Relatively poor performance is seen for GLM, ANN, and ARIMA. The performance deterioration can be due to lack of ability to capture complex spatio-temporal dependencies from data. Furthermore, smaller sample size is limiting ARIMA to capture correlation from such noisy time-series data. It is evident that for all metrics, the FOCIR-Net outperforms all the benchmark machine learning models using both datasets. For demand forecasting, the FOCIR-Net has 11.32 % and 4.38 % lower RMSE than the best benchmark GBM for Beijing and Hangzhou, respectively. Furthermore, the FOCIR improves the MAE by 6.25 % and 4.21 % than the best benchmarks XGBoost for Beijing and GBM for Hangzhou, respectively. For supply-demand gap forecasting, 9.73 % and 11.92 % improvement in RMSE and MAE, respectively, are seen for the FOCIR-Net than the best benchmarks GBM and XGBoost in Beijing, while 4.58 % improvement in MAE and marginal improvement of RMSE is found for FOCIR-Net than the best benchmark XGBoost in Hangzhou. However, the FOCIR-Net has more than 6 % improvement of sMAPE than the best benchmark XGBoost for supply-demand gap forecasting in both datasets.

Table 3 presents the results of the model ablation for the FOCIR-Net components. For both datasets, improvements around 2 % and 1 % in the RMSE of the FOCIR-Net are observed with respect to the corresponding values of RMSE in the OCIR-Net and the FOC-Net, respectively, which indicates that the feature importance layer and zone-distributed IndRNN adds to the performance of the FOCIR-Net. The marginal improvement of the FOCIR-Net over the FOC-Net may be due to the fact that the IndRNN requires more training data like the other recurrent neural networks (e.g. LSTM) for getting significant improvement [54]. Furthermore, the differences in temporal dependencies among different zones make the learning process of the IndRNN more difficult without the 1D CNN. This is more evident from the large drop in performance of the FIR-Net and zone-distributed IndRNN, which indicates that the 1D CNN plays the most important role in the FOCIR-Net.

Table 4 presents the results of the feature ablation for different combinations of features. Large deterioration of performance due to the removal of the spatio-temporal and context features indicates their utmost importance, whereas small performance drop due to the removal of temporal features indicates their trivial importance.

## 5.3 | Model interpretations

In this paper, the weights utilized by the feature importance layer in the FOCIR-Net are processed through a sigmoid function, which bounds the outputs of the learned weights between 0 and 1. Furthermore, weights in the feature importance layer are trained with L1 regularization and subsequent layers are trained with L2 regularization. Since sparsity is induced only in the feature importance layer and weights cannot become zero in the subsequent layers, output of the sigmoid activation for a feature in the feature importance layer indicates the contribution of that feature to the prediction and is termed as the importance score of that feature. The importance scores of the features can be explained in a similar manner as the co-efficients of the linear models, i.e. larger absolute values indicate greater contribution to prediction. In order to separately explain the contribution of the features temporally and spatially, the normalized outputs of the feature importance scores are averaged spatially and temporally, respectively. For conciseness, only the results for Hangzhou data are presented here.

Figures 7(a) and 7(b) presents the feature importance rankings of the spatially averaged features for demand and supply-demand gap forecasting, respectively. It is intriguing to find that the historical values of the spatio-temporal features, i.e. demand, quantity supplied and congestion have more contribution to the prediction than the temporal and context features. The least contribution among the spatio-temporal features for demand forecasting is seen for the historical values of the supply-demand gap, which may be due to the fact that supply-demand gap has irregular patterns whereas demand has recurring patterns. The same reason might be responsible for the lower contribution of the temporal and context features in supply-demand gap prediction than that of in demand prediction. It is also found that almost all the context features except POI and sleep hour are found to have more contribution towards prediction while comparing with the temporal features. Among the temporal features, weather, temperature and PM2.5 for the interval (t-6) have relatively higher importance than the other temporal features.

The temporally averaged features are presented in the heatmaps in Figures 8(a,b). In general, it is found that the features of the higher demand (supply-demand gap) zones are more important to the FOCIR-Net for prediction than that of the lower demand (supply-demand gap) zones. Furthermore, it is found that importance of the spatio-temporal features and the context features are not uniform across the zones, while importance of the temporal features are mostly uniform across the zones except for few zones (e.g. 8, 9, 16, 21, 23, 32, 38, 44), which are among the top zones with higher total orders as well as supply-demand gap as found in Figure 5(b). This clearly shows the importance of the spatial weighting in the feature importance layer: although values of the temporal features in a time-slot are same across all zones, their effects might not be same in all zones. It is also evident that the spatio-temporal and temporal features are more important than the temporal features for many of the zones. This finding is consistent with the results of the feature ablation, while clearly showing the advantage of feature importance layer for automatically emphasizing the features according to their importance.

## 6 | CONCLUSIONS

In this paper, a spatio-temporal deep learning architecture, FOCIR-Net, is proposed for forecasting both demand and supply-demand gap in a ride-hailing system with anonymised spatial adjacency information. The proposed architecture integrates feature importance layer with a spatio-temporal deep learning architecture composed of 1D CNN and zone-distributed IndRNN. The architecture of FOCIR-Net processes different types of spatio-temporal, temporal and context features by assigning spatial weights through the feature importance layer and learning spatial and temporal dependencies from the corresponding features through the 1D CNN and zone-distributed IndRNN. The weights learned from the feature importance layer further assists in the ranking of the features and interpreting the model. The proposed FOCIR-Net is compared against several benchmark models including time-series models such as ARIMA, and machine learning algorithms such as XGBoost, GBM, RF, XRT, GLM, and ANN. The models are tested with two real-world datasets of Didi from Beijing and Hangzhou, which shows the superiority of FOCIR-Net in terms of MAE, RMSE, and sMAPE in both demand and supply-demand forecasting tasks. An ablation analysis of the FOCIR-Net is conducted, which shows that the processing of the spatio-temporal variables through the 1D CNN is the

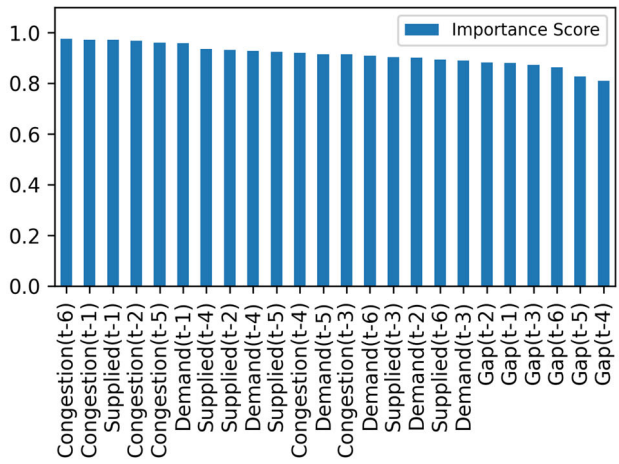
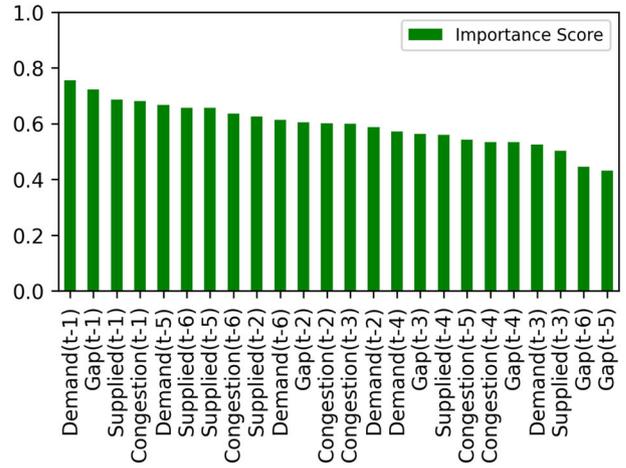
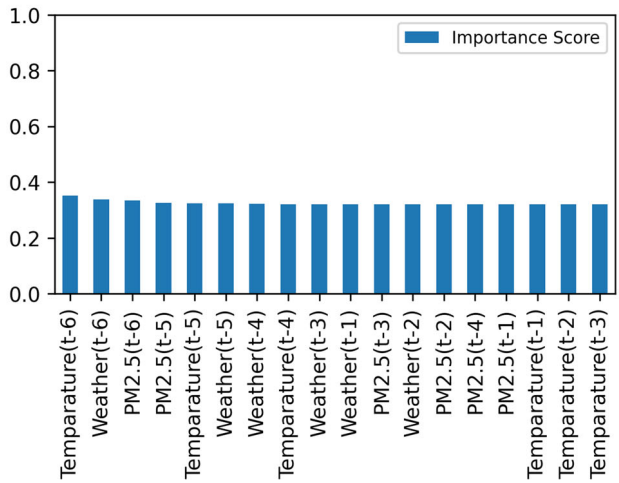
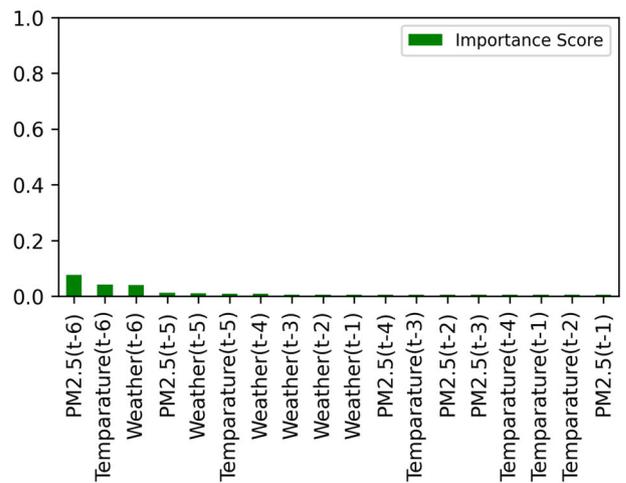
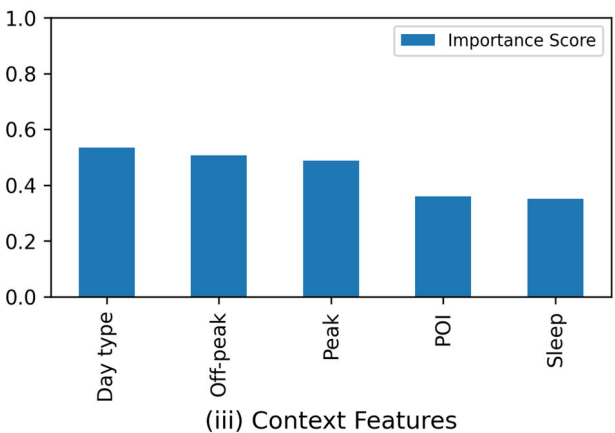
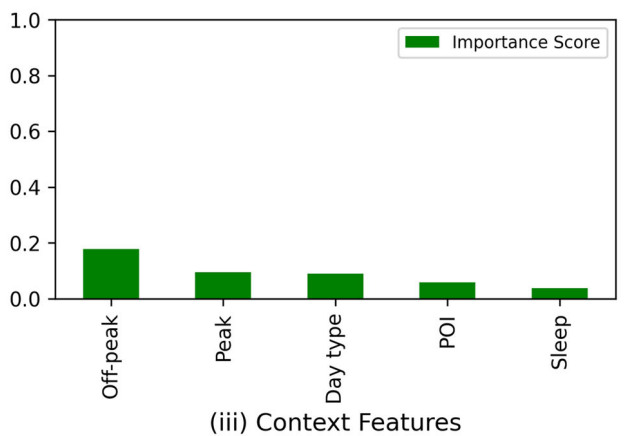

(a) Feature importance ranking for demand forecasting

(b) Feature importance ranking for gap forecasting

**FIGURE 7** Spatially averaged feature importance

most crucial part of the FOCIR-Net. Finally, interpretations of the models are provided based on the outputs of the feature importance layer, i.e. feature importance scores. From the spatially averaged importance scores it is found that the spatio-temporal features have the highest contribution to the prediction of both demand and supply-demand gap. Furthermore, the temporal and context features are more important for demand forecasting than that of for supply-demand gap forecasting.

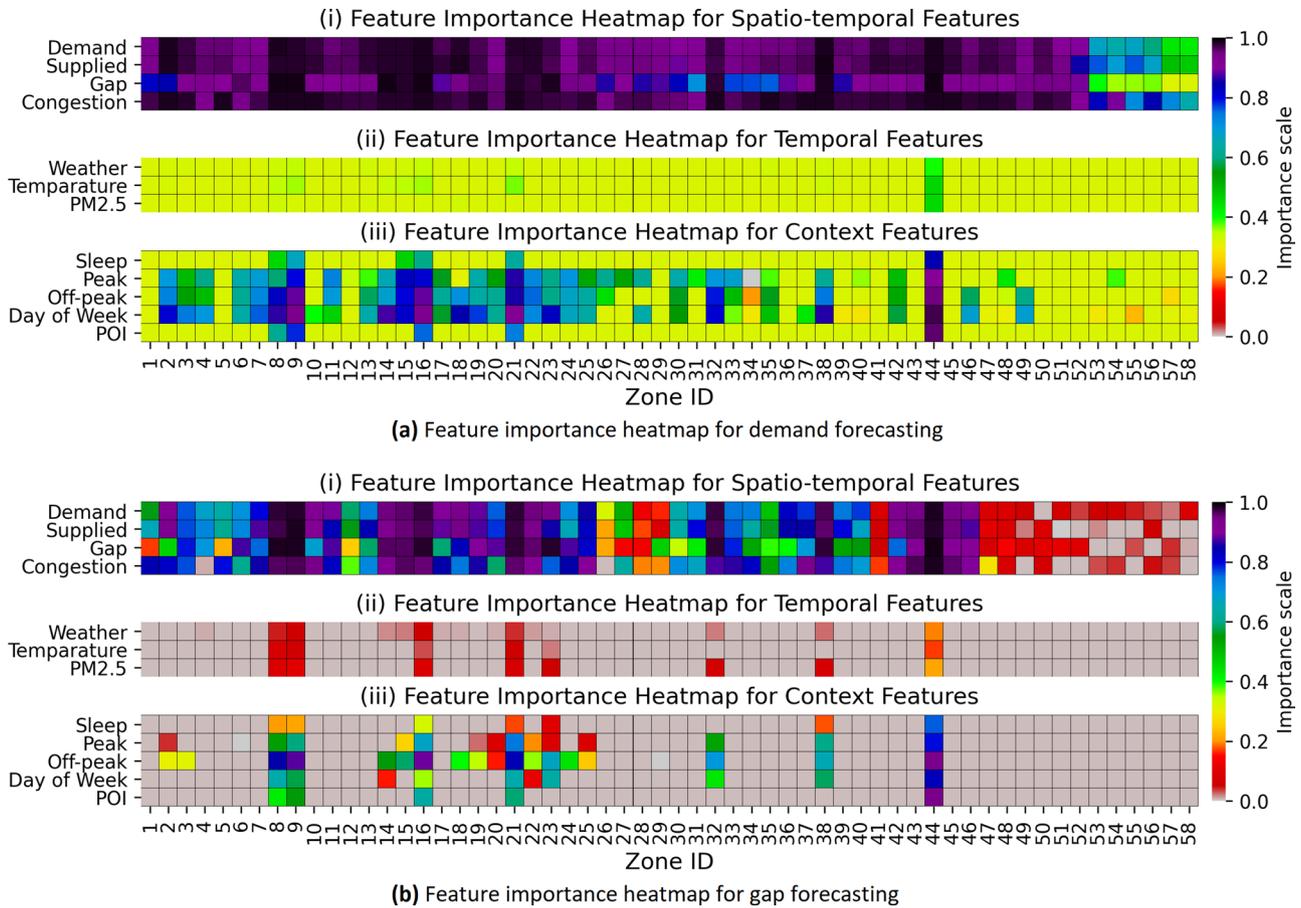

**FIGURE 8** Temporally averaged feature importance

From the temporally averaged importance scores it is seen that the effect of spatio-temporal and context features are more non-uniform across the zones for supply-demand gap forecasting than that of for demand forecasting. Furthermore, the effects of temporal features are mostly uniform with some exceptions, which cannot be accounted for without the spatial weighting in the feature importance layer.

The proposed architecture demonstrates the applicability of spatio-temporal deep learning for forecasting with anonymised spatial adjacency information. Furthermore, model interpretation shows that the feature importance layer can assist the researchers to get a better understanding of the spatio-temporal deep learning models. Nevertheless, our paper is not above limitations. Detailed information for some of the features such as weather categories, traffic congestion levels, and POI are unavailable in the datasets due to confidentiality issues that limited us from utilizing more features. The proposed architecture will be tested against a large number of features in the future by extending our architecture for station-based platforms such as taxi and bike-sharing. Furthermore, model interpretations show generalizations of some of the features between the demand and supply-demand gap models, which indicate that developing a multi-task learning architecture for simultaneous forecasting of demand and supply-demand gap can be an interesting area for future research.

**ACKNOWLEDGMENTS**

This research is funded by Miyan Research Institute, International University of Business Agriculture and Technology. The authors are thankful to Didi Chuxing for the publicly released datasets.

**FUNDING INFORMATION**

Miyan Research Institute, International University of Business Agriculture and Technology.

**ORCID**

*Md. Hishamur Rahman*  https://orcid.org/0000-0001-7555-5703